\documentclass[runningheads]{llncs}
\usepackage[T1]{fontenc}
\usepackage{graphicx}
\usepackage{hyperref}
\usepackage{xcolor}
\hypersetup{
    colorlinks,
    linkcolor={blue!50!black},
    citecolor={blue!50!black},
    urlcolor={blue!50!black}
}
\usepackage{color}
\usepackage{bbm}
\usepackage{amsmath}
\usepackage{amssymb}
\usepackage{graphicx,lipsum}

\begin{document}

% ---- Title ----
\title{B-Cos Aligned Transformers Learn Human-Interpretable Features}
\titlerunning{B-Cos Transformer}

% ---- Authors ----
\author{
Manuel Tran \inst{1, 2, 3} \and
Amal Lahiani \inst{1} \and
Yashin Dicente Cid \inst{1} \and
Melanie Boxberg \inst{3, 5} \and
Peter Lienemann \inst{2, 4} \and
Christian Matek \inst{2, 6} \and
Sophia J. Wagner \inst{2, 3} \and
Fabian J. Theis \inst{2, 3} \and
Eldad Klaiman \inst{1,}\thanks{Equal contribution.} \and
Tingying Peng \inst{2, \ast} 
}

\authorrunning{M. Tran et al.}

% ---- Institutes ----
\institute{
Roche Diagnostics / Penzberg / Germany \and
Helmholtz AI, Helmholtz Munich / Neuherberg / Germany \and
Technical University of Munich / Munich / Germany \and
Ludwig Maximilian University of Munich / Munich / Germany \and
Pathology Munich-North / Munich / Germany \and
University Hospital Erlangen / Erlangen / Germany
}

\let\oldmaketitle\maketitle
\renewcommand{\maketitle}{\oldmaketitle\setcounter{footnote}{0}}
\maketitle              % typeset the header of the contribution
\begin{sloppypar} 

% ---- Abstract ----
\begin{abstract}
Vision Transformers (ViTs) and Swin Transformers (Swin) are currently state-of-the-art in computational pathology. However, domain experts are still reluctant to use these models due to their lack of interpretability. This is not surprising, as critical decisions need to be transparent and understandable. The most common approach to understanding transformers is to visualize their attention. However, attention maps of ViTs are often fragmented, leading to unsatisfactory explanations. Here, we introduce a novel architecture called the B-cos Vision Transformer (BvT) that is designed to be more interpretable. It replaces all linear transformations with the B-cos transform to promote weight-input alignment. In a blinded study, medical experts clearly ranked BvTs above ViTs, suggesting that our network is better at capturing biomedically relevant structures. This is also true for the B-cos Swin Transformer (Bwin). Compared to the Swin Transformer, it even improves the F1-score by up to 4.7\% on two public datasets.


\end{abstract}

% ---- Introduction ----
\section{Introduction}
\label{introduction}

\begin{figure*}[!t]
\centering
    \includegraphics[width=0.8\textwidth]{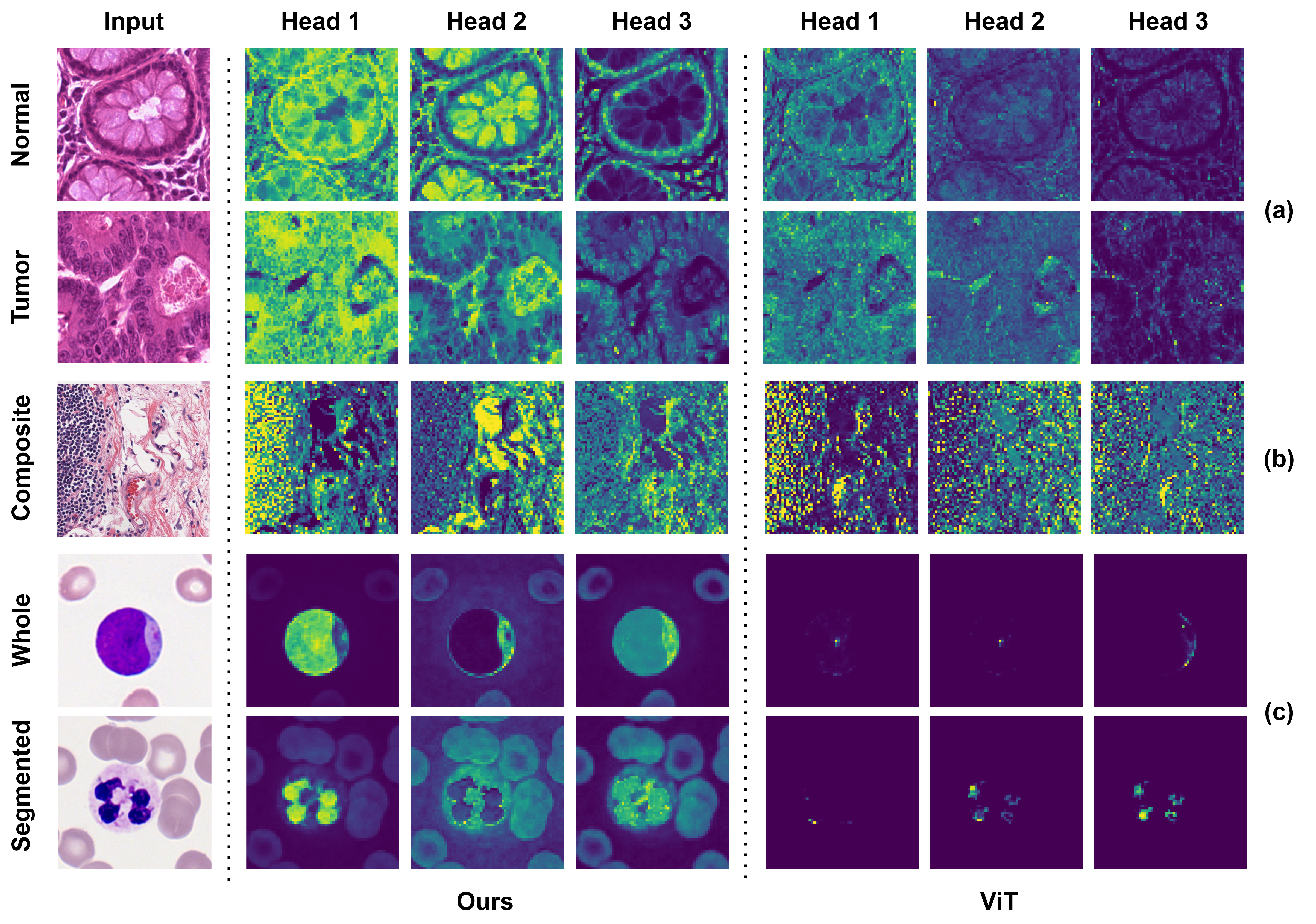}
    \caption{Attention maps of ViT and BvT (ours) on the test set of (a) NCT-CRC-HE-100K, (b) TCGA-COAD-20X, and (c) Munich-AML-Morphology. BvT attends to various diagnostically relevant features such as cancer tissue, cells, and nuclei.}
    \label{fig:overview}
\end{figure*}

Making artificial neural networks more interpretable, transparent, and trustworthy remains one of the biggest challenges in deep learning. They are often still considered black boxes, limiting their application in safety-critical domains such as healthcare. Histopathology is a prime example of this. For years, the number of pathologists has been decreasing while their workload has been increasing~\cite{Markl_2021_Virchows}. Consequently, the need for explainable computer-aided diagnostic tools has become more urgent.

As a result, research in explainable artificial intelligence is thriving~\cite{Linardatos_Entropy_2020}. Much of it focuses on convolutional neural networks (CNNs)~\cite{Du_ACM_2020}. However, with the rise of transformers~\cite{Vaswani_NIPS_2017} in computational pathology, and their increasing application to cancer classification, segmentation, survival prediction, and mutation detection tasks~\cite{Matsoukas_CVPR_2022,Wang_2022_MIA,Wagner_2023_arXiv}, the old tools need to be reconsidered. Visualizing filter maps does not work for transformers, and Grad-CAM~\cite{Selvaraju_CVPR_2017} has known limitations for both CNNs and transformers. 

The usual way to interpret transformer-based models is to plot their multi-head self-attention scores~\cite{Caron_2021_ICCV}. But these often lead to fragmented and unsatisfactory explanations~\cite{Chefer_CVPR_2021}. In addition, there is an ongoing controversy about their trustworthiness~\cite{Bibal_2022_ACL}. To address these issues, we propose a novel family of transformer architectures based on the B-cos transform originally developed for CNNs~\cite{Bohle_CVPR_2022}. By aligning the inputs and weights during training, the models are implicitly forced to learn more biomedically relevant and meaningful features (Figure~\ref{fig:overview}). Overall, our contributions are as follows: 

\begin{itemize}
    \item We propose the B-cos Vision Transformer (BvT) as a more explainable alternative to the Vision Transformer (ViT)~\cite{Dosovitskiy_2021_ICLR}.

    \item We extensively evaluate both models on three public datasets: NCT-CRC-HE-100K~\cite{Kather_2019_PLOS}, TCGA-COAD-20X~\cite{Kirk_COAD_2016}, Munich-AML-Morphology~\cite{Matek_NatMachIntell_2019}.

    \item We apply various post-hoc visualization techniques and conduct a blind study with domain experts to assess model interpretability.

    \item We derive the B-cos Swin Transformer (Bwin) based on the Swin Transformer~\cite{Liu_2021_ICCV} (Swin) in a generalization study.
\end{itemize}

% ---- Related Work ----
\section{Related Work}
\label{sec:relatedwork}

\begin{figure*}[!t]
\centering
    \includegraphics[width=0.9\textwidth]{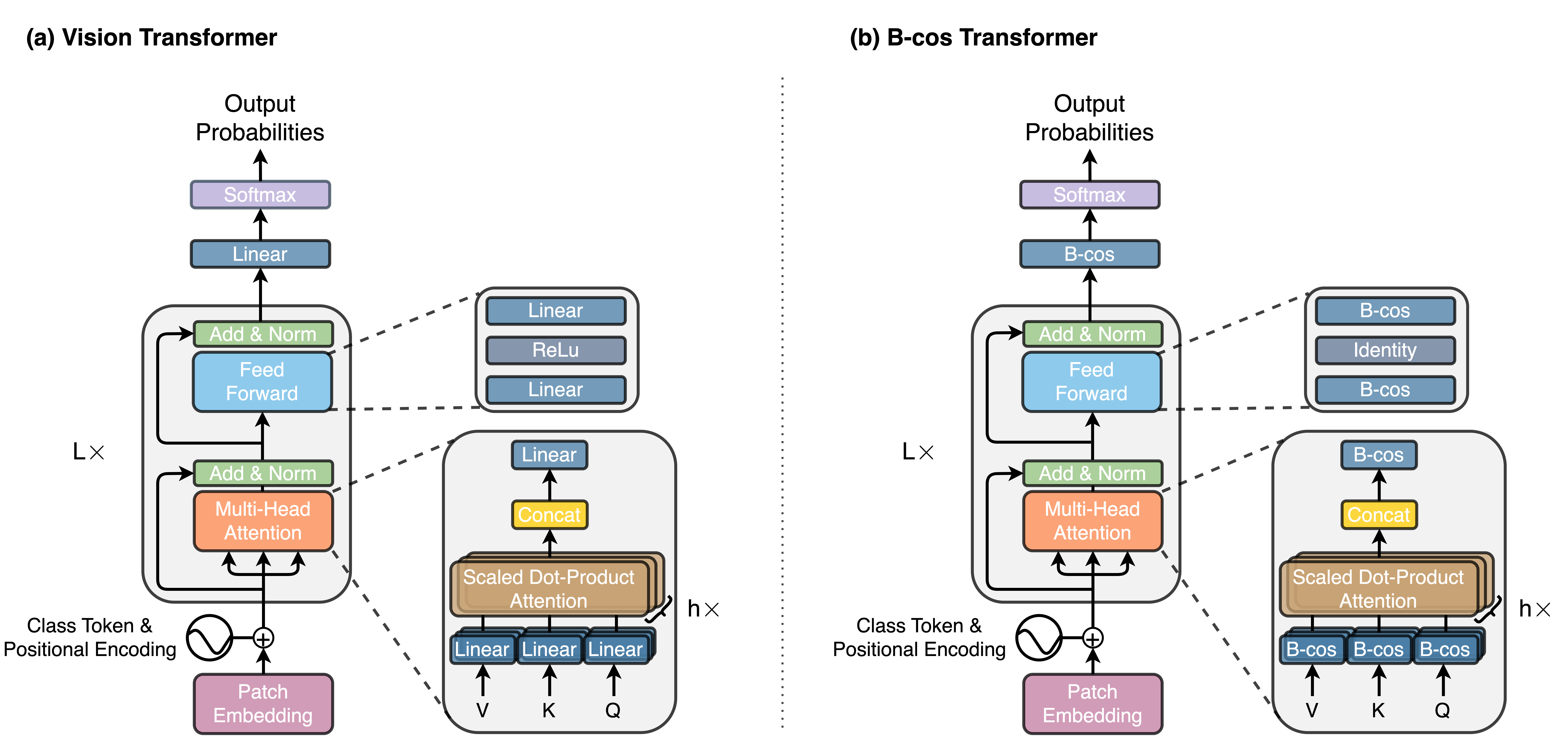}
    \caption{The model architecture of ViT and BvT (ours). We replace all linear transformations in ViT with the B-cos transform and remove all ReLU activation functions.}
    \label{fig:methods}
\end{figure*}

Explainability, interpretability, and relevancy are terms used to describe the ability of machine learning models to provide insight into their decision-making process. Although these terms have subtle differences, they are often used interchangeably in the literature~\cite{Gilpin_2018_DSAA}. 

Recent research on understanding vision models has mostly focused on attribution methods~\cite{Linardatos_Entropy_2020,Du_ACM_2020}, which aim to identify important parts of an image and highlight them in a saliency map. Gradient-based approaches like Grad-CAM~\cite{Selvaraju_CVPR_2017} or attribution propagation strategies such as Deep Taylor Decomposition~\cite{Montavon_Pattern_2017} and LRP~\cite{Binder_ICANN_2016} are commonly used methods. Perturbation-based techniques, such as SHAP~\cite{Lundberg_NeurIPS_2017}, are another way to extract salient features from images. Besides saliency maps, one can also visualize the activations of the model using Activation Maximization~\cite{Erhan_UOM_2009}.

However, it is still controversial whether the above methods can correctly reflect the behavior of the model and accurately explain the learned function (model-faithfulness~\cite{Jacovi_2020_ACL}). For example, it has been shown that some saliency maps are independent of both the data on which the model was trained and the model parameters~\cite{Adebayo_NeurIPS_2018}. In addition, they are often considered unreliable for medical applications~\cite{Arun_RAI_2021}. As a result, inherently interpretable models have been proposed as a more reliable and transparent solution. The most recent contribution are B-cos CNNs~\cite{Bohle_CVPR_2022}, which use a novel nonlinear transformation (the B-cos transformation) instead of the traditional linear transformation. 

\begin{figure*}[!t]
\centering
    \includegraphics[width=0.85\textwidth]{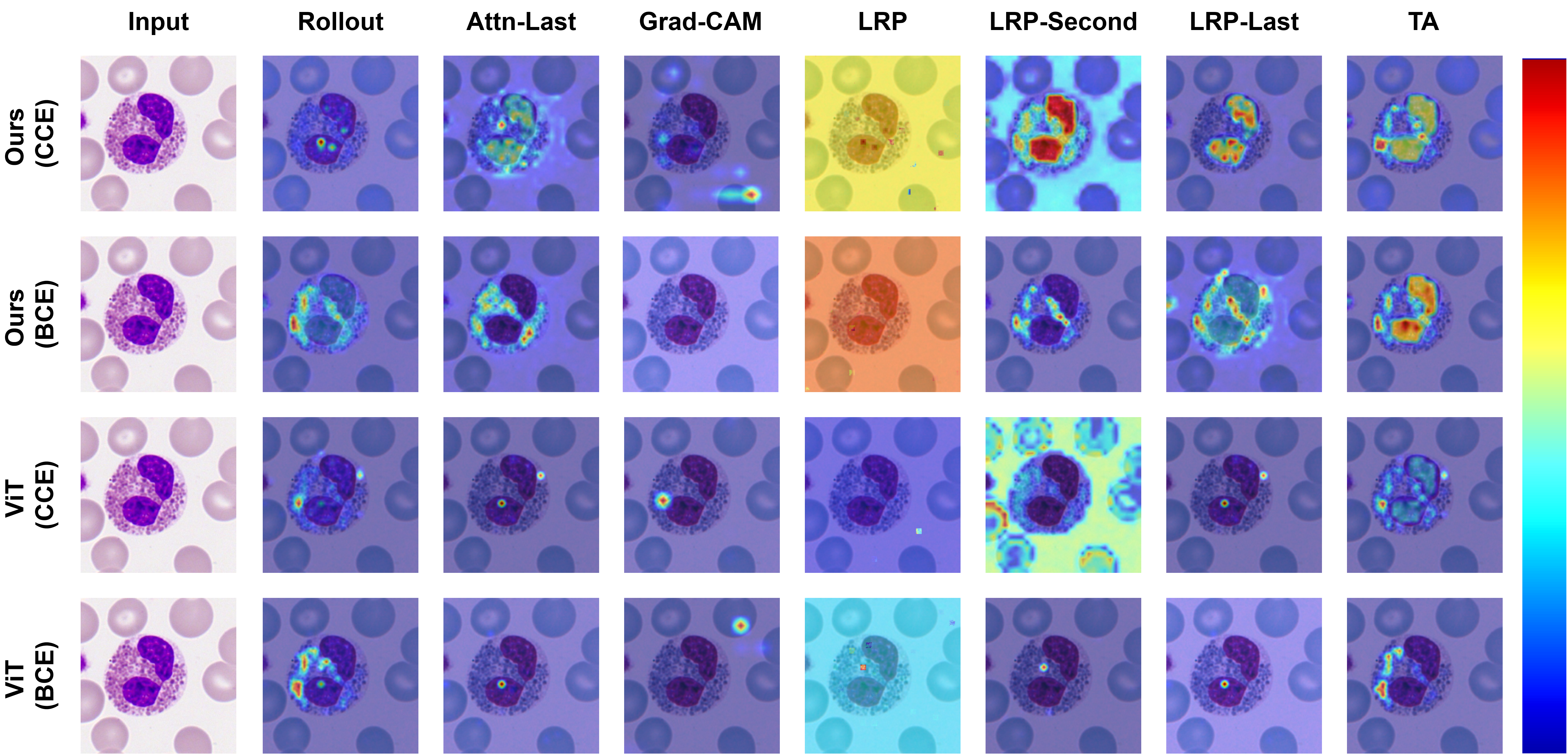}
    \caption{Rollout, Attention-Last (Attn-Last), Grad-CAM, LRP, LRP of the second layer (LRP-Second), LRP of the last layer (LRP-Last), and Transformer Attribution (TA) applied on the test set of Munich-AML-Morphology. The image shows an eosinophil, which is characterized by its split, but connected nucleus, large specific granules (pink structures in the cytoplasm), and dense chromatin (dark spots inside the nuclei)~\cite{Rosenberg_2013_NatRevImm}. Across all visualization techniques, BvT focuses on these exact features unlike ViT.} 
    \label{fig:aml}
\end{figure*}

Compared to CNNs, there is limited research on understanding transformers beyond attention visualization~\cite{Chefer_CVPR_2021}. Post-hoc methods such as Grad-CAM~\cite{Selvaraju_CVPR_2017} and Activation Maximization~\cite{Erhan_UOM_2009} used for CNNs can also be applied to transformers. But in practice, the focus is on visualizing the raw attention values (see Attention-Last~\cite{Hollenstein_ACL_2021}, Integrated Attention Maps~\cite{Dosovitskiy_2021_ICLR}, Rollout~\cite{Abnar_ACL_2020}, or Attention Flow~\cite{Abnar_ACL_2020}). More recent approaches such as Generic Attention~\cite{Chefer_CVPR_2021_b}, Transformer Attribution~\cite{Chefer_CVPR_2021}, and Conservative Propagation~\cite{Ali_ICMl_2022} go a step further and introduce novel visualization techniques that better integrate the attention modules with contributions from different parts of the network. Note that these methods are all post-processing methods applied after training to visualize the network's reasoning. On the other hand, the ConceptTransformer~\cite{Rigotti_2022_ICLR}, achieves better explainability by cross-attending user-defined concept tokens in the classifier head during training. Unlike all of these methods, interpretability is already an integral part of our architecture. Therefore, these methods can be easily applied to our models as we also show in our evaluations (Figure~\ref{fig:aml} and Figure~\ref{fig:swin}).

% ---- Methods ----
\section{Methods}
\label{methods}

We focus on the original Vision Transformer~\cite{Dosovitskiy_2021_ICLR}: The input image is divided into non-overlapping patches, flattened, and projected into a latent space of dimension $d$. Class tokens [\textit{cls}] are then prepended to these patch embeddings. In addition, positional encodings [\textit{pos}] are added to preserve topological information. In the scaled dot-product attention~\cite{Vaswani_NIPS_2017}, the model learns different features (query $Q$, key $K$, and value $V$) from the input vectors through a linear transformation. Both query and key are then correlated with a scaled dot-product and normalized with a softmax. These self-attention scores are then used to weight the value by importance:
\begin{equation}
  \text{Attention}(Q,K,V) = \text{softmax}\left( \dfrac{Q K^T}{\sqrt{d}} \right) V.
  \label{eq:attn}
\end{equation}

To extract more information, this process is repeated $h$ times in parallel (multi-headed self-attention). Each self-attention layer is followed by a fully-connected layer consisting of two linear transformations and a ReLU activation. 

We propose to replace all linear transforms in the original ViT (Figure~\ref{fig:methods})
\begin{equation}
  \text{Linear}(x,w) = w^Tx = \|w\|\|x\| c(x,w),
  \label{eq:linear}
\end{equation}
\begin{equation}
   c(x,w) = \text{cos}(\angle{(x,w)}), \quad \angle{} ... \text{angle between vectors}
  \label{eq:cos}
\end{equation}
with the B-cos* transform~\cite{Bohle_CVPR_2022}
\begin{equation}
  \text{B-cos*}(x;w) = \underbrace{\|\hat{w} \|}_{= 1} \|x\| |c(x, \hat{w})|^B \times \text{sgn} (c(x, \hat{w})),
  \label{eq:bcos*}
\end{equation}
where $B \in \mathbb{N}$. Similar to~\cite{Bohle_CVPR_2022}, an additional nonlinearity is applied after each B-cos* transform. Specifically, each input is processed by two B-cos* transforms, and the subsequent MaxOut activation passes only the larger output. This ensures that only weight vectors with higher cosine similarity to the inputs are selected, which further increases the alignment pressure during optimization. Thus, the final B-cos transform is given by
\begin{equation}
  \text{B-cos}(x;w) = \max_{i \in \{1,2\}}{\text{B-cos*}(x;w_i)}.
  \label{eq:bcos}
\end{equation}

\begin{figure*}[!t]
\centering
    \includegraphics[width=0.85\textwidth]{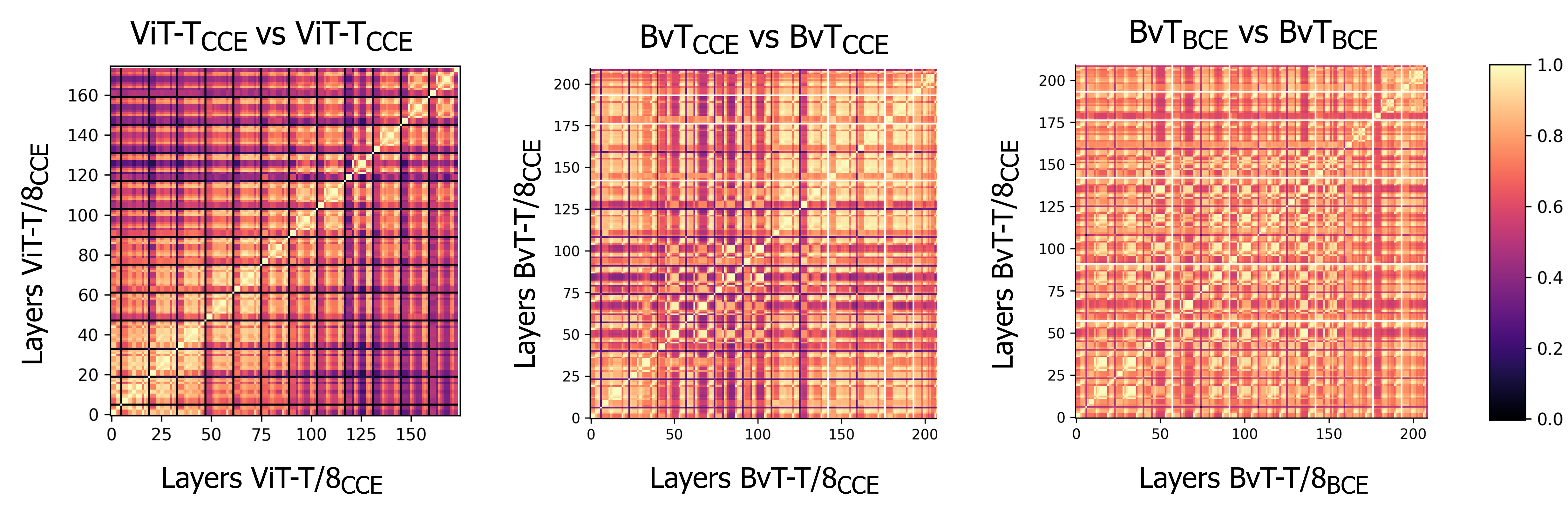}
    \caption{We compute the central kernel alignment (CKA), which measures the representation similarity between each hidden layer. Since the B-cos transform aligns the weights with the inputs, BvT (ours) achieves a more uniform representation structure compared to ViT (values closer to 1). When trained with the binary cross-entropy loss (BCE) instead of the categorical cross-entropy loss (CCE), the alignment is higher.}
    \label{fig:cka}
\end{figure*}

To see the significance of these changes, we look at Equation~\ref{eq:bcos*} and derive
\begin{equation}
  \|\hat{w} \| = 1 \Rightarrow \text{B-cos*}(x;w) \le \|x\|.
  \label{eq:allign}
 \end{equation}
Since $|c(x, \hat{w})| \le 1$, equality is only achieved if $x$ and $w$ are collinear, i.e., if they are aligned. Intuitively, this forces the weight vector to be more similar to the input. Query, key, and value thus capture more patterns in an image -- which the attention mechanism can then attend to. This can be shown visually by plotting the centered kernel alignment (CKA). It measures the similarity between layers by comparing their internal representation structure. Compared to ViTs, BvTs achieve a highly uniform representation across all layers (Figure~\ref{fig:cka}).

% ---- Implementation and Evaluation Details ----
\section{Implementation and Evaluation Details}
\label{sec:implementation}

\hspace{\parindent}\textbf{Task-based evaluation:} Cancer classification and segmentation is an important first step for many downstream tasks such as grading or staging. Therefore, we choose this problem as our target. We classify image patches from the public colorectal cancer dataset NCT-CRC-HE-100K~\cite{Kather_2019_PLOS}. We then apply our method to TCGA-COAD-20X~\cite{Kirk_COAD_2016}, which consists of 38 annotated slides from the TCGA colorectal cancer cohort, to evaluate the effectiveness of transfer learning. This dataset is highly unbalanced and not color normalized compared to the first dataset. Additionally, we demonstrate that the B-cos Vision Transformer is adaptable to domains beyond histopathology by training the model on the single white blood cell dataset Munich-AML-Morphology~\cite{Matek_NatMachIntell_2019}, which is also highly unbalanced and also publicly available.

\textbf{Domain-expert evaluation:} Our primary objective is to develop an extension of the Vision Transformer that is more transparent and trusted by medical professionals. To assess this, we propose a blinded study with four steps: (i) randomly selecting images from the test set of TCGA-COAD-20X (32 samples) and Munich-AML-Morphology (56 samples), (ii) plotting the last-layer attention and transformer attributions for each image, (iii) anonymizing and randomly shuffling the outputs, (iv) submitting them to two domain experts in histology and cytology for evaluation. 

\begin{figure*}[!t]
\centering
    \includegraphics[width=1.0\textwidth]{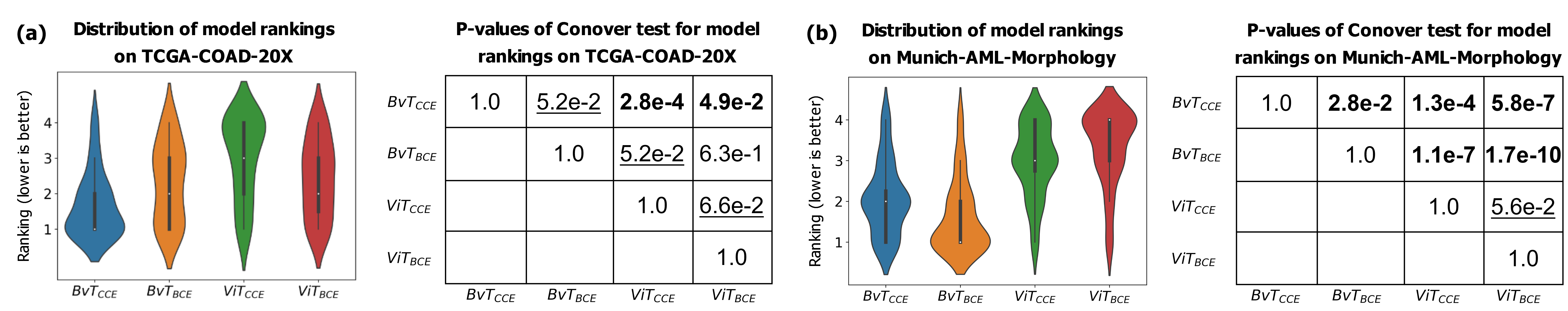}
    \caption{In a blinded study, domain experts ranked models (lower is better) based on whether the models focus on biomedically relevant features that are known in the literature to be important for diagnosis. We then performed the Conover post-hoc test after Friedman with adjusted p-values according to the two-stage Benjamini-Hochberg procedure. BvT ranks above ViT with $p$$<$$0.1$ (underlined) and $p$$<$$0.05$ (bold).}
    \label{fig:analysis}
\end{figure*}

\begin{table}[!b]
\centering
\caption{F1-score, top-1, and top-3 accuracy from the test set of NCT-CRC-HE-100K, Munich-AML-Morphology, and TCGA-COAD-20X. We compare ViT and BvT (ours) trained with categorical cross-entropy (CCE) and binary cross-entropy (BCE) loss using two model configurations: T/8 and S/8 (see Section \ref{sec:implementation}).}
\label{tab:vit}
\begin{tabular}{l|ccc|ccc|ccc}
\hline
 & \multicolumn{3}{c|}{\textbf{NCT}} & \multicolumn{3}{c|}{\textbf{Munich}} & \multicolumn{3}{c}{\textbf{TCGA}} \\ 
\textbf{Models} & \textbf{F1} & \textbf{Top-1} & \textbf{Top-3} & \textbf{F1} & \textbf{Top-1} & \textbf{Top-3} & \textbf{F1} & \textbf{Top-1} & \textbf{Top-3} \\ \hline & \\[-1.0em]
$\text{ViT}$-T/8$_\text{CCE}$ & \textbf{90.9} & 92.7 & 99.1 & \textbf{57.3} & 90.1 & 98.9 & 57.1 & 78.8 & 94.4 \\
$\text{ViT}$-S/8$_\text{CCE}$ & 89.2 & 91.1 & 99.5 & 56.3 & 93.1 & 99.0 & 56.3 & 78.6 & 92.9 \\ \hline & \\[-1.0em]
$\text{BvT}$-T/8$_\text{CCE}$ & 88.8 & 91.1 & 99.3 & 54.0 & 87.1 & 98.6 & \textbf{61.0} & 77.4 & 93.1 \\
$\text{BvT}$-S/8$_\text{CCE}$ & 88.4 & 90.1 & 99.4 & 52.9 & 89.8 & 98.6 & 60.2 & 76.3 & 92.9 \\\hline \hline & \\[-1.0em]
$\text{ViT}$-T/8$_\text{BCE}$ & 90.0 & 91.4 & 98.4 & 54.8 & 90.0 & 99.0 & 53.6 & 79.6 & 93.9 \\
$\text{ViT}$-S/8$_\text{BCE}$ & \textbf{90.2} & 92.2 & 99.3 & \textbf{55.4} & 92.8 & 99.0 & 54.1 & 77.0 & 88.9 \\ \hline & \\[-1.0em]
$\text{BvT}$-T/8$_\text{BCE}$ & 86.7 & 90.1 & 98.5 & 51.1 & 83.5 & 97.9 & 57.7 & 79.8 & 93.4 \\
$\text{BvT}$-S/8$_\text{BCE}$ & 87.5 & 90.4 & 99.4 & 52.4 & 85.0 & 98.3 & \textbf{59.0} & 74.5 & 88.9 \\ \hline %& \\[-1.0em]
\end{tabular}
\end{table}

\textbf{Implementation details:} In our experiments, we compare different variants of the B-cos Vision Transformer and the Vision Transformer. Specifically, we implement two versions of ViT: ViT-T/8 and ViT-S/8. They only differ in parameter size (5M for T models and 22M for S models) and use the same patch size of 8. All BvT models (BvT-T/8 and BvT-S/8) are derivatives of the corresponding ViT models. The B-cos transform used in the BvT models has an exponent of $B=2$. We use AdamW with a cosine learning rate scheduler for optimization and a separate validation set for hyperparameter selection. Following the findings of~\cite{Bohle_CVPR_2022}, we add $[1-r, 1-g, 1-b]$ to the RGB channels $[r, g, b]$ of BvT. This allows us to encode each pixel with the direction of the color channel vector, forcing the model to capture more color information. Furthermore, we train models with two different loss functions: the standard categorical cross-entropy loss (CCE) and the binary cross-entropy loss (BCE) with one-hot encoded entries. It was suggested in~\cite{Bohle_CVPR_2022} that BCE is a more appropriate loss for B-cos CNNs. We explore whether this is also true for transformers in our experiments. Additional details on training, optimization, and datasets can be found in the Appendix.

% ---- Results and Discussion ----
\section{Results and Discussion}
\label{results}

\begin{figure*}[!t]
\centering
    % trim={<left> <lower> <right> <upper>}
    \includegraphics[width=0.75\textwidth, trim={0 1.5cm 0 1.5cm}, clip]{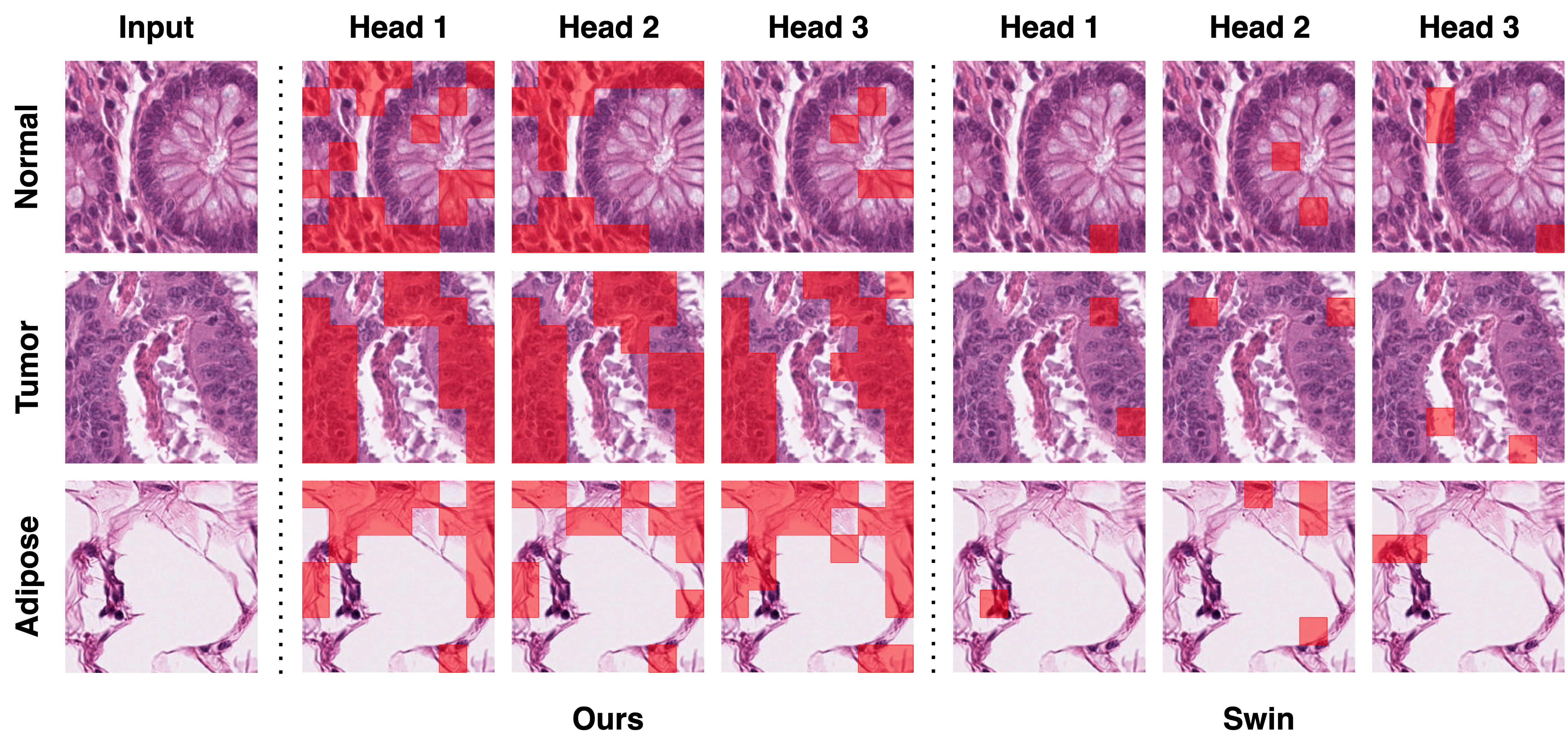}
    \caption{Attention maps of the last layer of the modified Swin and Bwin (ours). Bwin focuses on cells and nuclei, while Swin mostly focuses on a few spots.}
    \label{fig:swin}
\end{figure*}

\hspace{\parindent}\textbf{Task-based evaluation:} When trained from scratch, all BvT models underperform their ViT counterparts by about 2\% on NCT-CRC-HE-100K and 3\% on Munich AML-Morphology (Table~\ref{tab:vit}). However, when we use the pre-trained weights from NCT-CRC-HE-100K and transfer them to TCGA-COAD-20X for fine-tuning, BvT outperforms ViT by up to 5\% (Table~\ref{tab:vit}). We believe this is due to the simultaneous optimization of two objectives: classification loss and weight-input alignment. With a pre-trained model, BvT is likely to focus more on the former. In addition, we observe that models trained with BCE tend to perform worse than those trained with CCE. However, their attention maps seem to be more interpretable (see Figure~\ref{fig:aml}).

\textbf{Domain-expert evaluation:} The results show that BvTs are significantly more trustworthy than ViTs ($p<0.05$). This indicates that BvT consistently attends to biomedically relevant features such as cancer cells, nuclei, cytoplasm, or membrane~\cite{Matek_Blood_2021} (Figure~\ref{fig:analysis}). In many visualization techniques, we see that BvT, unlike ViT, focuses exclusively on these structures (Figure~\ref{fig:aml}). In contrast, ViT attributes high attention to seemingly irrelevant features, such as the edges of the cells. A third expert points out that ViT might overfit certain patterns in this dataset, which could aid the model in improving its performance.

% ---- Generalization Study ----
\section{Generalization to Other Architectures}
\label{generalization}

\begin{table}[!t]
\centering
\caption{Results of Swin and Bwin (ours) experiments on the test set of NCT-CRC-HE-100K and Munich-AML-Morphology. We report F1-score, top-1, and top-3 accuracy.}
\label{tab:swin}
\begin{tabular}{l|ccc|ccc}
\hline
\textbf{} & \multicolumn{3}{c|}{\textbf{NCT}} & \multicolumn{3}{c}{\textbf{AML}} \\
\textbf{Models} & \textbf{F1} & \textbf{Top-1} & \textbf{Top-3} & \textbf{F1} & \textbf{Top-1} & \textbf{Top-3} \\ \hline
$\text{Swin}$-T$_\text{CCE}$ & 89.1 & 92.1 & 99.0 & 48.2 & 94.1 & 98.8 \\
%$\text{Swin}$-T$_\text{CCE}$ (modified first layer) & 89.3 & 92.4 & 99.4 & 50.1 & 94.0 & 98.8 \\
$\text{Swin}$-T$_\text{CCE}$ (modified) & 89.8 & 92.0 & 99.6 & 49.1 & 94.2 & 98.9 \\ \hline & \\[-1.0em]
$\text{Bwin}$-T$_\text{CCE}$ & 91.5 & 93.5 & 99.5 & 53.0 & 93.9 & 98.6 \\
%$\text{Bwin}$-T$_\text{CCE}$ (modified first layer) & 92.2 & \textbf{94.4} & 99.8 & \textbf{56.7} & 92.6 & 98.7 \\
$\text{Bwin}$-T$_\text{CCE}$ (modified) & \textbf{92.5} & 94.3 & 99.6 & \textbf{53.3} & 93.8 & 98.7 \\ \hline
\end{tabular}
\end{table}

We aim to explore whether the B-cos transform can enhance the interpretability of other transformer-based architectures. The Swin Transformer (Swin)~\cite{Liu_2021_ICCV} is a popular alternative to ViT (e.g., it is currently the SOTA feature extractor for histopathological images~\cite{Wang_2022_MIA}). Swin utilizes window attention and feed-forward layers. In this study, we replace all its linear transforms with the B-cos transform, resulting in the B-cos Swin Transformer (Bwin). However, unlike BvT and ViT, it is not obvious how to visualize the window attention. Therefore, we introduce a modified variant here that has a regular ViT / BvT block in the last layer.

In our experiments (Table~\ref{tab:swin}), we observe that Bwin outperforms Swin by up to 2.7\% and 4.8\% in F1-score on NCT-CRC-HE-100K and Munich-AML-Morphology, respectively. This could be attributed to the window attention, which reintroduces many of the inductive biases of CNNs. Moreover, we would like to emphasize that the modified models have no negative impact on the model's performance. In fact, all metrics remain similar or even improve. The accumulated attention heads (we keep 50\% of the mass) demonstrate that Bwin solely focuses on nuclei and other cellular features (Figure~\ref{fig:swin}). Conversely, Swin has very sparse attention heads, pointing to a few spots. Consistent with the BvT vs ViT blind study, our pathologists also agree that Bwin is more plausible than Swin ($p<0.05$). Additional details can be found in the Appendix.

% ---- Conclusion ----
\section{Conclusion}
\label{conclusion}

We have introduced the B-cos Vision Transformer (BvT) and the B-cos Swin Transformer (Bwin) as two alternatives to the Vision Transformer (ViT) and the Swin Transformer (Swin) that are more interpretable and explainable. These models use the B-cos transform to enforce similarity between weights and inputs. In a blinded study, domain experts clearly preferred both BvT and Bwin over ViT and Swin. We have also shown that BvT is competitive with ViT in terms of quantitative performance. Moreover, using Bwin or transfer learning for BvT, we can even outperform the original models in downstream classification tasks.

\iffalse
\begin{table*}[!htbp]
\centering
\begin{tabular}{lccccc}
\hline
& \multicolumn{2}{c}{\textbf{Pascal VOC-2012}} & \multicolumn{3}{c}{\textbf{Oxford-IIIT Pets}} \\ \hline
& \textbf{CIFAR-10 \phantom{}} & \textbf{\phantom{} ImageNette \phantom{}} 
& \multicolumn{1}{l}{\textbf{\phantom{} CIFAR-10 \phantom{}}} & \multicolumn{1}{l}{\textbf{\phantom{} ImageNette \phantom{}}} & \textbf{\phantom{} ImageWoof} \\ \hline
\multicolumn{1}{l|}{BvT-T/8} & \textbf{45.8} & \multicolumn{1}{c|}{\textbf{44.5}} & \textbf{49.5} & \textbf{59.6} & \textbf{60.6} \\
\multicolumn{1}{l|}{ViT-T/8} & 18.8 & \multicolumn{1}{c|}{20.3} & 20.3 & 24.4 & 19.3 \\ \hline & \\[-1.0em]
\multicolumn{1}{l|}{DINO-S/8} & \multicolumn{2}{c|}{41.8} & \multicolumn{3}{c}{42.5} \\ \hline
\end{tabular}
\caption{We use the attention heads of BvT-T/8 and ViT-T/8 (trained on CIFAR-10, ImageNette, or ImageWoof) as segmentation masks on Pascal VOC-2012 and Oxford-IIIT Pets and compare the IoU to those of ViT-S/8 pre-trained with DINO (ImageNet).}
\label{tab:segmentation}
\end{table*}
\fi

%Please place your acknowledgments at the end of the paper, preceded by an unnumbered %run-in heading (i.e. 3rd-level heading).

\bibliographystyle{splncs04}
\bibliography{legacy}

\end{sloppypar} 
\end{document}

% ---- Title ----
\title{Appendix: B-Cos Aligned Transformers Learn Human-Interpretable Features}
\titlerunning{Appendix: B-Cos Transformer}

% ---- Authors ----
\author{
Manuel Tran \inst{1, 2, 3} \and
Amal Lahiani \inst{1} \and
Yashin Dicente Cid \inst{1} \and
Melanie Boxberg \inst{3, 5} \and
Peter Lienemann \inst{2, 4} \and
Christian Matek \inst{2, 6} \and
Sophia J. Wagner \inst{2, 3} \and
Fabian J. Theis \inst{2, 3} \and
Eldad Klaiman \inst{1,}\thanks{Equal contribution.} \and
Tingying Peng \inst{2, \ast} 
}

\authorrunning{M. Tran et al.}

% ---- Institutes ----
\institute{
Roche Diagnostics / Penzberg / Germany \and
Helmholtz AI, Helmholtz Munich / Neuherberg / Germany \and
Technical University of Munich / Munich / Germany \and
Ludwig Maximilian University of Munich / Munich / Germany \and
Pathology Munich-North / Munich / Germany \and
University Hospital Erlangen / Erlangen / Germany
}

\let\oldmaketitle\maketitle
\renewcommand{\maketitle}{\oldmaketitle\setcounter{footnote}{0}}
\maketitle              % typeset the header of the contribution
\begin{sloppypar}

% ---- Dataset ----
\section{Dataset}
\label{sec:dataset}

We evaluated our method on three public datasets: NCT-CRC-HE-100K, TCGA-COAD-20X, and Munich-AML-Morphology. Additional information about the datasets can be found in Table~\ref{tab:data}:

\begin{table}[]
\centering
\caption{We list details about the dataset used: number of training, validation, and test samples; class imbalance ratio; patch size in pixels; and resolution in microns per pixel. After tuning the hyperparameters, we merged the training and validation sets.}
\label{tab:data}
\begin{tabular}{l|cccccc}
 & \textbf{num\_train} & \textbf{num\_val} & \textbf{num\_test} & \textbf{cls\_imb} & \textbf{patch\_size} & \textbf{resolution} \\ \\ \hline 
\textbf{NCT}  & 90,000 & 10,000 & 7,180  & 1.6 & 224 px & 0.50 mpp \\
\textbf{AML}  & 11,025 & 3,666  & 3,674  & 771 & 400 px & 0.07 mpp \\
\textbf{TCGA} & 78,653 & 22,712 & 24,411 & 861 & 512 px & 0.50 mpp \\ \hline 
\end{tabular}
\end{table}

% ---- Training ----
\section{Training}
\label{sec:training}

We trained ViT and BvT on NCT-CRC-HE-100K for 10 epochs. We then fine-tuned the models for two epochs on TCGA-COAD-20X. On Munich-AML-Morphology, we performed a warm restart after 15 epochs, decreased the learning rate by a factor of 10, and trained the models for another 15 epochs (except for BvT-S/8, which was trained for another 30 epochs). Both Swin and Bwin were trained for 6 epochs on NCT-CRC-HE-100K. Similar to before, we also performed a warm restart on Munich-AML-Morphology after 30 epochs. We restarted with a learning rate 10 times smaller than the initial one and trained for another 15 epochs. For training, we used an NVIDIA Tesla V100-SXM2-32GB. Each model required less than 12h to train. Total emissions were estimated to be 91.5kg CO$_2$. Estimations were conducted using the \href{https://mlco2.github.io/impact#compute}{MachineLearning Impact calculator} presented in~\cite{Lacoste_arXiv_2019}. More training hyperparameters are listed in Table~\ref{tab:hyper}. 

\begin{table}[]
\centering
\caption{List of additional hyperparameters for all models.}
\label{tab:hyper}
\begin{tabular}{l|cccccc}
 & \textbf{optimizer} & \textbf{scheduler} & \textbf{learning\_rate} & \multicolumn{1}{l}{\textbf{weight\_decay}} & \textbf{batch\_size} \\ \\ \hline 
\textbf{ViT-T/8} & AdamW & cosine & 1e-4 & 1e-4 & 64 \\
\textbf{ViT-S/8} & AdamW & cosine & 1e-4 & 1e-4 & 16 \\
\textbf{BvT-T/8} & AdamW & cosine & 1e-3 & 1e-4 & 64 \\
\textbf{BvT-S/8} & AdamW & cosine & 1e-3 & 1e-4 & 16 \\
\textbf{Swin-T}  & AdamW & cosine & 1e-4 & 1e-4 & 32 \\
\textbf{Bwin-T}  & AdamW & cosine & 1e-3 & 1e-4 & 32 \\ \hline 
\end{tabular}
\end{table}

% ---- Blinded Study ----
\section{Blinded Study}
\label{sec:study}

\begin{figure*}[]
\centering
    \includegraphics[width=0.5\textwidth]{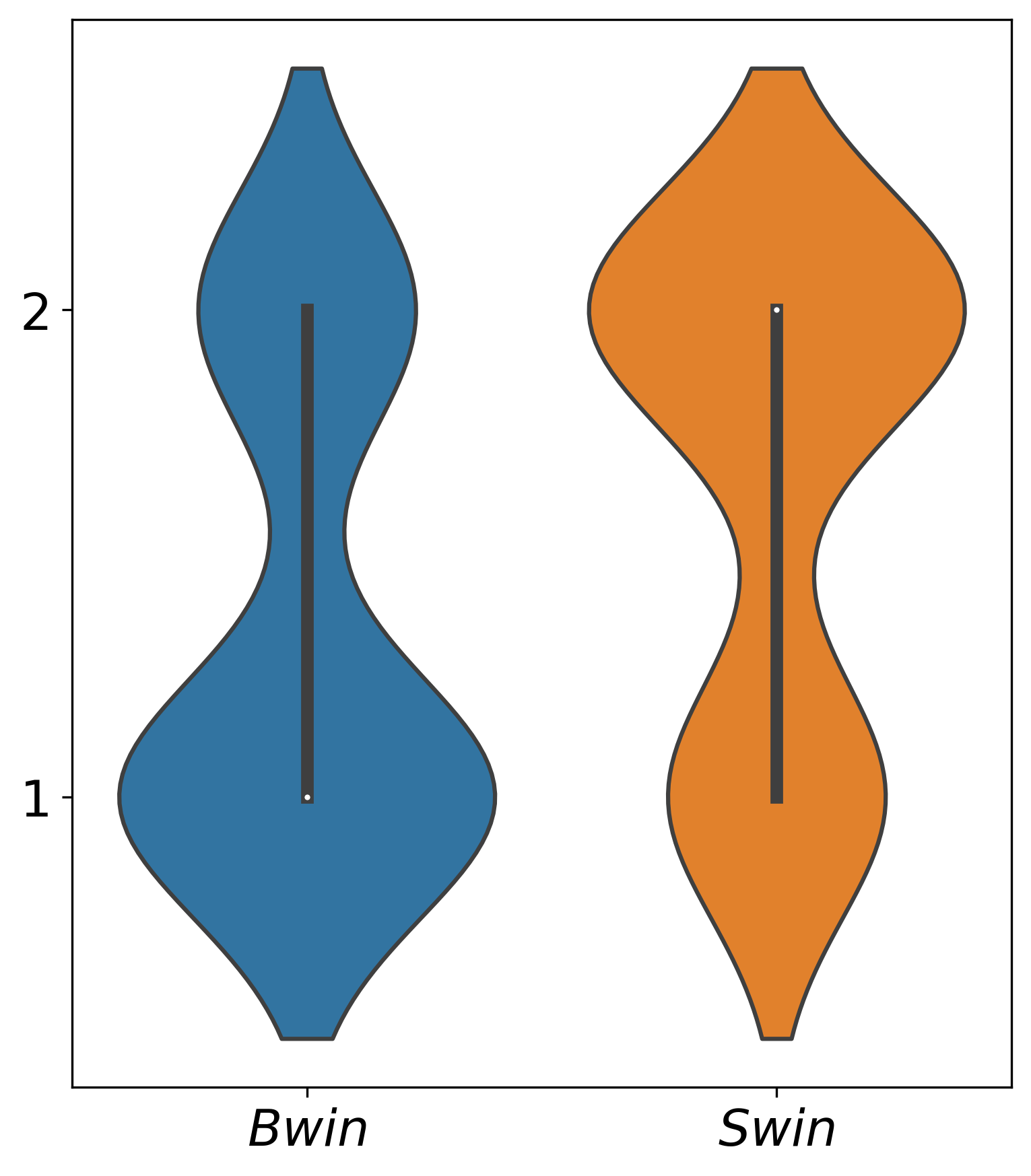}
    \caption{In a blinded study, a domain expert ranked models (lower is better) based on whether the models focus on biomedically relevant features that are known in the literature to be important for diagnosis. We then performed the Student t-test, showing that Bwin is better than Swin ($p<0.05$).}
    \label{fig:study}
\end{figure*}

We conducted another blinded study with 30 randomly selected images from NCT-CR-CHE-100K for the Swin and Bwin experiments, ensuring that each of the 9 classes was represented at least three times in the test. We then plotted the final attention maps from the modified Swin and Bwin models and randomly shuffled the anonymized outputs. To compare performance, we used the Student t-test (Figure~\ref{fig:study})

\bibliographystyle{splncs04}
\bibliography{legacy}

\end{sloppypar}